\documentclass[a4paper,11pt,article,oneside]{memoir}
\usepackage{ecis2015}

\usepackage{xspace}
\newcommand\ie{i.\,e.\xspace}
\newcommand\eg{e.\,g.\xspace}

\usepackage{csquotes}
\usepackage{amsmath}

\usepackage{xcolor}

\usepackage{siunitx}
    \DeclareSIUnit\eur{\officialeuro}

\usepackage{amssymb}
\usepackage{bm}

\usepackage{eurosym}

\usepackage{cleveref}
  \crefname{chapter}{Section}{Sections}
  \Crefname{chapter}{Section}{Sections}

\maintitle{Improving Decision Analytics with Deep Learning: The Case of Financial Disclosures} 
\shorttitle{Deep Learning for Financial Disclosures} 
\category{Research in Progress} 

\authors{
Feuerriegel, Stefan, University of Freiburg, Freiburg, Germany, stefan.feuerriegel@is.uni-freiburg.de

Fehrer, Ralph, University of Freiburg, Freiburg, Germany, ralphfehrer@gmail.com}
\shortauthors{Feuerriegel and Fehrer} 

\bibliography{literature} 

\begin{document}

\begin{abstract}
Decision analytics commonly focuses on the text mining of financial news sources in order to provide managerial decision support and to predict stock market movements. Existing predictive frameworks almost exclusively apply traditional machine learning methods, whereas recent research indicates that these methods are not sufficiently capable of extracting suitable features and capturing the non-linear nature of complex tasks. As a remedy, novel deep learning models aim to overcome this issue by extending classical neural networks with additional hidden layers. Indeed, deep learning often provides a viable approach to achieve a high predictive performance. In this paper, we adapt the novel deep learning technique to financial decision support, where we aim to predict the direction of stock movements following financial disclosures. As a result, our paper shows how deep learning can outperform the accuracy of benchmarks for machine learning by 5.66\,\%.
\end{abstract}



\chapter{Introduction}

Organizations are constantly looking for ways to improve their decision-making processes in core areas, such as marketing, firm communication, production and procurement \citep{Turban.2011}. While the classical approach relies on having humans devise simple decision-making rules, modern decision support can also exploit statistical evidence that originates from analyzing data \citep{Apte.2002,Arnott.2005,AsadiSomeh.2015,Boylan.2012,Davenport.2006,Vizecky.2011b}. This data-driven decision support is nowadays also fueled by the \emph{Big Data} era \citep{Boyd.2012,Chen.2012b,Halper.2011,Power.2014}. The term Big Data usually refers to data that is massive in size. In addition, such data comes often in different formats (\eg video, text), changes quickly and is subject to uncertainty \citep{IBM.2013}.

Crucial aspects of data-driven decision support systems entail the prediction of future events, such as consumer behavior or stock market reactions to press releases, based on an analysis of historical data \citep{Apte.2002,Vizecky.2011b}. Decision analytics thus frequently utilizes modeling, machine learning and data mining techniques from the area of \emph{predictive analytics}. In fact, predictive analytics can be instrumented for \emph{\textquote{generating new theory, developing new measures, comparing competing theories, improving existing theories, assessing the relevance of theories, and assessing the predictability of empirical phenomena}} \citep{Shmueli.2011}.

Predictive analytics frequently contributes to managerial decision support, as is the case when predicting investor reaction to press releases and financial disclosures \citep{Nassirtoussi.2014}. In this instance, predictive analytics is typically confronted with massive datasets of heterogeneous and mostly textual content, while outcomes are simultaneously of high impact for any business. Until now, decision support for financial news still predominantly relies on traditional machine learning techniques, such as support vector machines or decision trees \citep{Minev.2012,Nassirtoussi.2014,Pang.2008}. 

The performance of traditional machine learning algorithms largely depends on the features extracted from underlying data sources, which has consequently elicited the development and evaluation of feature engineering techniques \citep{Arel.2010}. Research efforts to automate and optimize the feature engineering process, along with a growing awareness of current neurological research, has led to the emergence of a new sub-field of machine learning research called \emph{deep learning} \citep{Hinton.2006}. Deep learning takes into account recent knowledge on the way the human brain processes data and thus enhances traditional neural networks by a series of hidden layers. This series of hidden layers allows for deeper knowledge representation, possibly resulting in improved predictive performance. Deep learning methods have been applied to well-known challenges in the machine learning discipline, such as pattern recognition and natural language processing. The corresponding results indicate that deep learning can outperform classical machine learning methods (which embody only a shallow knowledge layer) in terms of accuracy \citep[c.\,f.][]{Hinton.2006,Socher.2011}.
 
In this paper, we want to unleash the predictive power of deep learning to provide decision-support in the financial domain. As a common challenge, we choose the task of predicting stock market movements that follow the release of a financial disclosure. We expect that deep learning can learn appropriate features from the underlying textual corpus efficiently and thus surpass other state-of-the-art classifiers. However, the successful application of deep learning techniques is not an easy task; deep learning implicitly performs feature extraction through the interplay of different hidden layers, the representation of the textual input and the interactions between layers. In order to master this challenge, we apply the recursive autoencoder model introduced by \citet{Socher.2011} and tailor it to the prediction of stock price directions based on the content of financial materials. 

The remainder of this paper is structured as follows. \Cref{sec:related_work} provides a short overview of related work in which we discuss similar text mining approaches and give an overview of relevant deep learning publications. We then explain our methodology and highlight the differences between classical and deep learning approaches (\Cref{sec:methodology}). Finally, \Cref{sec:evaluation} evaluates both approaches using financial news disclosures and discuss the managerial implications. 

\chapter{Related Work}
\label{sec:related_work}

This Information Systems research is positioned at the intersection between finance, Big Data, decision support and predictive analytics. The first part of this section discusses traditional approaches of providing decision support based on financial news. In the second part, we discuss previous work that focuses on the novel deep learning approach.

\section{Decision analytics for financial news}

Text mining of financial disclosures represents one of the fundamental approaches for decision analytics in the finance domain. The available work can be categorized by the necessary \emph{preprocessing} steps, the \emph{text mining algorithms}, the underlying \emph{text source} (\eg press releases, financial news, tweets) and its \emph{focus} on facts or opinions (\eg quarterly reports, analyst recommendations). While \citet{Pang.2008} provide a comprehensive domain-independent survey, other overviews concentrate solely on the financial domain \citep{Minev.2012,Mittermayer.2006b}. In a very recent survey, \citet{Nassirtoussi.2014} focus specifically on studies aimed at stock market prediction. We structure the discussion of the related research according to the above categories.

Among the most popular \emph{text mining algorithms} are classical machine learning algorithms, such as support vector machines, regression algorithms, decision trees and Na\"{i}ve Bayes. In addition, neural network models have been used more rarely, but are slowly gaining traction, just as in other application domains \citep{Nassirtoussi.2014}. Furthermore, Bayesian learning can provide explanatory insights by generating domain-dependent dictionaries \citep{Proellochs.2015}.

As part of \emph{preprocessing}, the first step in most text mining approaches is the generation of a set of values that represent relevant textual features, which can be used as inputs for the subsequent mining algorithms. This usually involves the selection of features based on the raw text sources, some kind of dimensionality reduction and the generation of a good feature representation, such as binary vectors. A comprehensive discussion of the various techniques used for feature engineering can be found in \citet{Nassirtoussi.2014,Pang.2008}.

The \emph{text sources} used for text mining include financial news \citep[e.\,g.][]{Alfano.2015,Feuerriegel.2013,Feuerriegel.2015,Feuerriegel.2016b} and company-specific disclosures, and range from the less formal, such as tweets \citep[e.\,g.][]{Bollen.2011}, to more formal texts, such as corporate filings \citep[e.\,g.][]{Feuerriegel.2016c,Muntermann.2007,Proellochs.2015}. Some researchers have focused exclusively on the headlines of news sources to exclude the noise usually contained in longer texts \citep{Peramunetilleke.2002}.

News disclosures with a fact-based \emph{focus} are especially relevant for investors. As such, German ad~hoc announcements in English contain strictly regulated content and a tight logical connection to the stock price, making them an intriguing application in research. The measurable effect of ad~hoc news on abnormal returns on the day of an announcement have been established by several authors \citep[c.\,f.][]{Groth.2011,Hagenau.2013,Muntermann.2007,Proellochs.2015}. Consequently, we utilize the same corpus in our following evaluation.


\section{Deep learning as an emerging trend}

Deep learning\footnote{For details, see reading list \emph{\textquote{Deep Learning}}. Retrieved April~21, 2015 from \url{http://deeplearning.net/reading-list/}.} originally focused on complex tasks, in which datasets are usually high-dimensional \citep{Arel.2010,Bengio.2009}. As such, one of the first successful deep learning architectures consisted of a so-called autoencoder in combination with a Boltzmann machine, where the autoencoder carries out unsupervised pre-training of the network weights \citep{Hinton.2006}. This model performs well on several benchmarks from the machine learning literature, such as image recognition. Moreover, its architecture can be adapted to enhance momentum stock trading \citep{Takeuchi.2013} as one of the few successful applications of deep learning to financial decision support. However, this publication relies only on past stock returns and neglects the predictive power of exogenous input, such as financial disclosures.

The natural language processing community has only recently started to adapt deep learning principles to the specific requirements of language recognition tasks. For example, \citet{Socher.2011} utilize a recursive autoencoder to predict sentiment labels based on individual movie review sentences. Further research improved the results on the same dataset by combining a recursive neural tensor network with a sentiment treebank \citep{Socher.2013}.


\chapter{Methodologies for Financial Decision Support}
\label{sec:methodology}

\begin{figure}
\includegraphics[width=\linewidth]{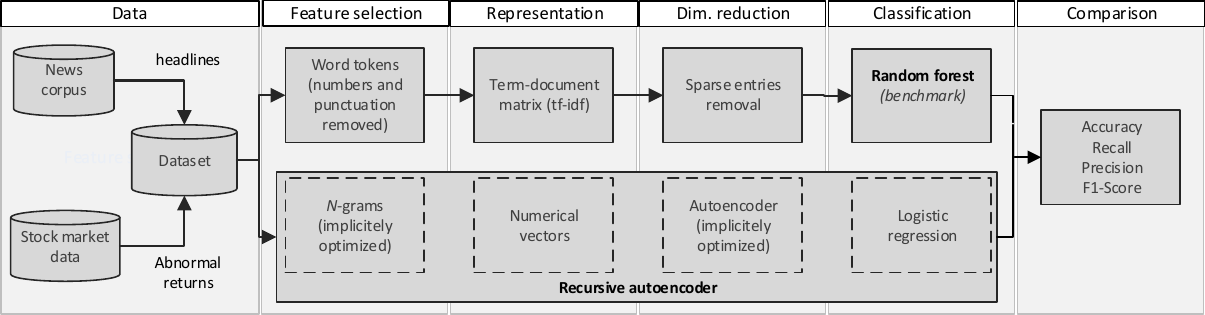}
\caption{Research framework comparing classical machine learning and deep learning.}
\label{fig:methodology}
\end{figure}

This section introduces our research framework to provide financial decision support based on news disclosures. In brevity, we introduce a benchmark classifier and our deep learning architecture to predict stock movements. Altogether, \Cref{fig:methodology} illustrates how we compare both prediction algorithms. The random forest and the recursive autoencoder are both trained to predict stock market directions based on the ad~hoc announcements and the according abnormal returns. To compare the performance of the recursive autoencoder to the benchmark, we apply the same test set to each of the trained algorithms and  measure the predictive performance in terms of accuracy, precision and recall based on the confusion matrix.

Following \citet{Nassirtoussi.2014}, we divide the overall procedure into steps for data generation, feature selection, feature reduction and feature representation. Both approaches, the benchmark algorithm and the recursive autoencoder, differ fundamentally in their preprocessing. The application of a random forest or support vector machine requires traditional feature engineering, whereas the recursive autoencoder, as a remedy, automatically generates a feature representation as part of its optimization algorithm. This is indicated in \Cref{fig:methodology} by the extension of the recursive autoencoder box over all preprocessing steps. 

\section{Benchmark: predicting stock movements with random forests}

In a first step, one transforms the running text into a matrix representation, which subsequently works as the input to the actual random forest algorithm. 
First of all, we remove numbers, punctuations and stop words from the running text and then split it into tokens \citep{Manning.1999}. Afterwards, we count the frequencies of how often terms occur in each news disclosure, remove sparse entries to reduce the dimensionality and store these values in a document-term matrix. The document-term matrix then represents the features. The actual values are weighted \citep{Salton.1983} by the \emph{term frequency-inverse document frequency} (tf-idf). This is a common approach in information retrieval to adjust the word frequencies by their importance.  

In the following evaluation, we utilize random forests as a benchmark classifier. Random forests represent one of the most popular machine learning algorithms due to their favorable predictive accuracy, relatively low computational requirements and robustness \citep{Breiman.2001,Hastie.2009,Kuhn.2013}. Random forests are an ensemble learning method for classification and regression, which is based on the construction and combination of many de-correlated decision trees. 

Given a training set $X = \left\{ \bm{x}_1, \ldots, \bm{x}_N \right\}$ with associated responses $Y= \left\{ y_1, \ldots, y_N\right\}$, the algorithm repeats the following steps $B$ times (we choose $B=500$): (1)~sample with replacement from $X$ and $Y$ to generate new subsets $X'$ and $Y'$. (2)~Train a decision tree $t_b$ using $X'$ and $Y'$. The individual decision trees $t_b$, $b = 1, \ldots, B$ can be combined to predict a response $\hat{y}$ for unseen values $\bm{x}$ as follows. One calculates individual predictions $t_b(\bm{x})$ for $b = 1, \ldots, B$ from each tree and then aggregates these predictions by simply using the majority vote to get the final response.

\section{Deep learning architecture: recursive autoencoders}

\begin{figure}
\includegraphics[width=.5\linewidth]{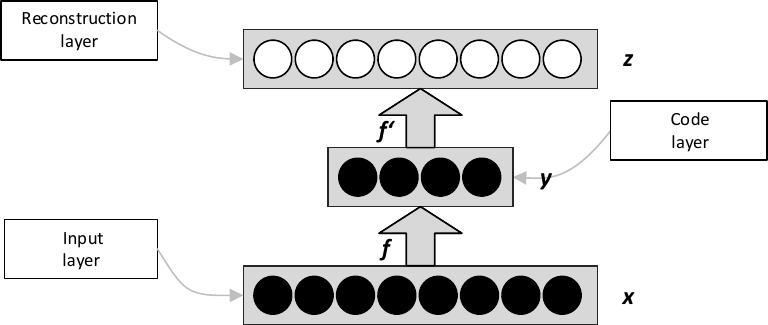}
\caption{An autoencoder searches a mapping between input $\bm{x}$ and a lower-dimensional representation $\bm{y}$ such that the reconstructed value $\bm{z}$ is similar to the input $\bm{x}$.}
\label{fig:autoencoder}
\end{figure}

This section describes the underlying architecture of our deep learning approach for financial disclosures based on so-called \emph{autoencoders}. The architecture of an autoencoder is illustrated in \Cref{fig:autoencoder}. An autoencoder is basically an artificial neural network, which finds a lower-dimensional representation of input values. Let $\bm{x}\in \left[ 0,1 \right]^N$ denote our input vector, for which we seek a lower-dimensional representation $\bm{y} \in \left[ 0,1 \right]^M$ with $M < N$. The mapping $f$ between $\bm{x}$ and $\bm{y}$ is named \emph{encoding} function and can be, generally speaking, any non-linear function, although a common choice is the sigmoid function 
\begin{equation}
f(\bm{x}) = \sigma(W \bm{x}+\bm{b})=\cfrac{1}{1+\exp{\left( W \bm{x}+\bm{b} \right)}} = \bm{y} \qquad\text{with parameters } W \text{ and } \bm{b}. 
\end{equation}
The key idea of an autoencoder is to find a second mapping from $\bm{y}$ to $\bm{z} \in \left[ 0,1 \right]^N$ given by $f'(\bm{y}) = \sigma(W' \bm{y}+\bm{b'})$, such that $\bm{z}$ is almost equal to the input $\bm{x}$. Mathematically speaking, we choose the free parameters in $f$ and $f'$ by minimizing the difference between the original input vector $\bm{x}$ and the reconstructed vector $\bm{z}$. This can be effectively achieved using numerical optimization, such as gradient descent, in order to determine the weights $W$ and $W'$. Altogether, the representation $\bm{y}$ (often called \emph{code}) is a lower-dimensional representation of the input data; it is frequently used as input features for subsequent learning because it only contains the most relevant or most discriminating features of the input space. 

\begin{figure}
\includegraphics[width=.75\linewidth]{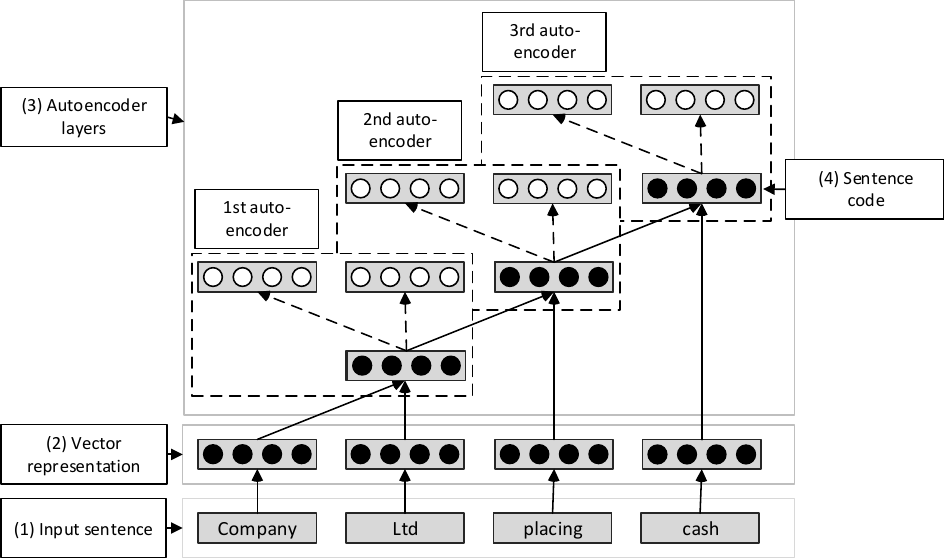}
\caption{A recursive autoencoder is a sequence of autoencoders, where each not only takes a vector as input but also the lower-dimensional code of the previous autoencoder in the sequence.}
\label{fig:recursive_autoencoder}
\end{figure}

The classical autoencoder works merely with a simple vector as input. In order to incorporate contextual information, we extend the classical autoencoder, resulting in a so-called \emph{recursive autoencoder}. Here, one trains a sequence of autoencoders, where each not only takes a vector $\bm{x}$ as input but also recursively the lower-dimensional code of the previous autoencoder in the sequence. Let us demonstrate this approach with an example as illustrated in \Cref{fig:recursive_autoencoder}. We process input in the form of a sequence of words. Each word is \eg given by a binary vector with zeros except for a single entry with $1$ representing the current word. Then, we train the first autoencoder with the input from the first two words \emph{Company} and \emph{Ltd}. Its lower-dimensional code is then input to the second autoencoder together with the vector representation of the word \emph{placing}. This recursion proceeds up to the final autoencoder, which produces as output the code representation for the complete sentence. Hence, this recursive approach aims to generate a compact code representation of a complete sentence while incorporating contextual information in the code layer. More precisely, this approach can learn from an ordered sequence of words and not only the pure frequencies. In addition, the recursive autoencoder entails an intriguing advantage: it can compress large input vectors in an unsupervised fashion without the need for class labels.

The basic steps for the recursive autoencoder are as follows: in a first step, the individual words of an input sentence (1) are mapped onto vectors of equal length $l$ (2). We initialize the values of the weights by sampling from a Gaussian distribution and later continuously updated through backpropagation. Through a recursive application of the autoencoder algorithm (3), the complete input sentence is then compressed bit by bit into a single code representation of length $l$ (4). For this purpose, the first autoencoder generates a code representation of length $l$ from the vectors representing the first two words in the sentence. The second autoencoder takes this code representation and the third word vector as inputs, and calculates the code representation of the next level. 


In order to extract and predict sentiment values, we use an extended variant of the recursive autoencoder model \citep{Socher.2011}, which includes an additional \emph{softmax layer} (sometimes also referred to as multinomial logit) in each autoencoder. This softmax function estimates the probability that an input vector $\bm{x}$ belongs to a certain class $j \in K$ via
\begin{equation}
P (y = j \,|\, \bm{x}) = \frac{\exp{\left( \bm{x}^{T} W_j \right)}}{\sum\limits_{k=1}^K \exp{\left( \bm{x}^{T} W_k \right)}} .
\end{equation}
In order to train this model, we optimize the weights $W_k$ of both the autoencoders and the softmax layers simultaneously with a combined target function. We then utilize the trained weights to classify unknown sentences by first computing the code representation inside the recursive autoencoder and, second, calculating the probabilities for each class from the softmax function. Interestingly, the backward mapping $f'$ is needed for training but is no longer needed for the prediction (that is why black circles indicate the vectors only necessary for prediction in \Cref{fig:autoencoder} and \Cref{fig:recursive_autoencoder}). 

\chapter{Evaluation: Predicting Stock Market Direction from Financial News}
\label{sec:evaluation}

In this section, we discuss and evaluate our experimental setting for predicting the direction of stock market movements following financial disclosures. We start with describing the steps involved in the generation of the underlying dataset and then compare classical machine learning with our deep learning architecture. 

\section{Dataset}

Our news corpus originates from regulated ad~hoc announcements\footnote{Kindly provided by Deutsche Gesellschaft f\"{u}r Ad-Hoc-Publizit\"{a}t~(DGAP).} between January~2004 and the end of June~2011 in English. These announcements conform to German regulations that require each listed company in Germany to immediately publish any information with a potentially significant effect on the stock price. With their direct relation to a particular stock, the tightly controlled content and the measurable effect on the stock price on the day of the announcement, ad~hoc announcements are particularly well-suited for the development and evaluation of techniques for predictive analytics.

Since recursive autoencoders work on sentence tokens, we exclusively use the headlines of English ad~hoc announcements for the prediction and discard the message body. As previous work \citep[e.\,g.][]{Peramunetilleke.2002} has shown, this is not a major disadvantage and can even help in reducing noise, as long as the titles concisely represent the content of the text.

We gather the financial data of the releasing companies from Thomson Reuters Datastream. We retrieve the firm performance with the help of the International Securities Identification Numbers~(ISIN) that appear first in each of the ad~hoc announcements. The stock price data before and on the day of the announcement are extracted using the corresponding trading day. These are then used to calculate abnormal returns \citep{MacKinlay.1997,Konchitchki.2011,Proellochs.2015}; abnormal returns can be regarded as some kind of excess return caused by the news release. In addition, we remove penny stocks with stock prices below \,\$5 for noise reduction. We then label each announcement title with one of three return direction classes (\emph{up}, \emph{down} or \emph{steady}), according to the abnormal return of the corresponding stock on the announcement day and discard the steady samples for noise reduction.

The resulting dataset consists of 8359 headlines from ad~hoc announcements with corresponding class labels \emph{up} or \emph{down}. Of this complete dataset, we use the samples covering the first 80\,\% of the timeline as training samples and the remaining 20\,\% as test samples.\footnote{We avoid the use of $k$-fold cross-validation as  this would neglect the timing of disclosures. For instance, we would evaluate our models with disclosures during the financial crisis, while the same models were previously trained with later knowledge of how news were perceived after the happening of the financial crisis.}

\section{Preliminary results}

We can now apply the above methods for predictive analytics to provide decision support regarding how investors react upon textual news disclosures. By comparing random forests and recursive autoencoders, we can evaluate our hypothesis that deep learning outperforms our benchmark in the current setting.


The detailed results are listed in \Cref{tbl:results}. We compare the predictive performance on the out-of-sample test set in terms of accuracy, precision, recall and the F1-score. The random forest as our benchmark achieves an in-sample accuracy of 0.63 and an out-of-sample accuracy of 0.53 at best. In comparison, the recursive autoencoder\footnote{We systematically tried several combinations for the two adjustable parameters embedding size (the length $l$ of the mapped feature vectors) and number-of-iterations (\ie number of gradient descent iterations). The best result accounts for an accuracy of 0.56 on the test-set, with a vector embedding size of~40 and 70 iterations. As expected, increasing the number of iterations usually results in better accuracy on the training set and lower accuracy on the test set -- a typical indication of over-fitting.} as our deep learning architecture results in an accuracy of 0.56. This accounts for a relative improvement of 5.66\,\%. Similarly, the F1-score increases from 0.52 to 0.56 -- a substantial rise of 7.69\,\%. The higher accuracy, as well as the improved F1-score, of the recursive autoencoder underlines our initial assumption that deep learning algorithms can outperform our benchmark from classical machine learning. When comparing the necessary computational resources, we see that recursive autoencoders ($\approx$\SI{23}{min}) require less computation time than decision trees (more than \SI{200}{min}).\footnote{Timings measured on an Intel Core i7-4700MQ CPU running at 2.4\,GHz with 8\,GB RAM and 64-bit Windows 8.1. Please note that programming languages and matrix algebra libraries vary which makes a fair comparison difficult.} Moreover, recursive autoencoders have an additional advantage: one can simply inject the complete set of news headlines as input without the manual effort of feature engineering. The reason for this is that the calculation and optimization of a feature representation is integrated into the optimization routines of deep learning algorithms.

The above results comply with the reported figures of around \SI{60}{\percent} with the full message body from related work \citep{Hagenau.2013,Groth.2008}. In direct comparison to the benchmarks, our evaluation provides evidence that deep learning is a compelling approach for the prediction of stock price movements. 

\begin{table}
\begin{tabular}{l SSSS} 
\toprule
\textbf{Predictive Analytics Method} & \textbf{Accuracy} &\textbf{Precision} &\textbf{Recall}  & \textbf{F1-Score} \\ 
\midrule
Random Forest & 0.53 & 0.53 & 0.51 & 0.52 \\
Recursive Autoencoder & 0.56  & 0.56 & 0.56  & 0.56 \\ 
\emph{Relative Improvement} & \SI{5.66}{\percent} & \SI{5.66}{\percent} & \SI{9.80}{\percent} & \SI{7.69}{\percent} \\ 
\bottomrule
\end{tabular}
\caption{Preliminary results evaluating improvements by utilizing deep learning to predict the direction of stock price movements following financial disclosures.}
\label{tbl:results}
\end{table}

\section{Discussion and implications for practitioners}

Traditional machine learning techniques still represent the default method of choice in predictive analytics. However, recent research indicates that these methods insufficiently capture the properties of complex, non-linear problems. Accordingly, the experiments in this paper show that a deep learning algorithm is capable of implicitly generating a favorable knowledge representation. 

As a recommendation to practitioners, better results are achievable with deep learning than with classical methods that rely on explicitly generated features. Nevertheless, practitioners must be aware of the complex architecture of deep learning models. This requires both a thorough understanding and solid experience in order to use such models efficiently. 

The economic impact of our improvement is manifold. A higher predictive performance enables business opportunities for automated traders. In addition, corporates can utilize this approach to assess the expected market response subsequent to disclosures. This works as a safety mechanism to check if the subjective perception of investors matches the content of a release.

\chapter{Conclusion and Research Outlook}

In the present paper, we show how a novel deep learning algorithm can be applied to provide decision support for the financial domain. Thereby, we contribute to Information Systems research by shedding light on how to exploit deep learning as a recent trend for managerial decision support. We demonstrate that a recursive autoencoder outperforms our benchmark from traditional machine learning in the prediction of stock market movements following financial news disclosures. The recursive autoencoder benefits from being able to automatically generate a deep knowledge representation.

In future research, we intend to broaden the preliminary results of this Research-in-Progress paper. First, our analysis could benefit substantially from incorporating and comparing further algorithms from predictive analytics. Second, we want to generalize our results by including further news sources.

\printbibliography

\end{document}